# Major Limitations of Satellite images

Firouz Abdullah Al-Wassai*
Research Student,
Computer Science Dept.
(SRTMU), Nanded, India
fairozwaseai@yahoo.com

N.V. Kalyankar
Principal,
Yeshwant Mahavidyala College
Nanded, India
drkalyankarnv@yahoo.com

*Abstract:* Remote sensing has proven to be a powerful tool for the monitoring of the Earth's surface to improve our perception of our surroundings has led to unprecedented developments in sensor and information technologies. However, technologies for effective use of the data and for extracting useful information from the data of Remote sensing are still very limited since no single sensor combines the optimal spectral, spatial and temporal resolution. This paper briefly reviews the limitations of satellite remote sensing. Also, reviews on the problems of image fusion techniques. The conclusion of this, According to literature, the remote sensing is still the lack of software tools for effective information extraction from remote sensing data. The trade-off in spectral and spatial resolution will remain and new advanced data fusion approaches are needed to make optimal use of remote sensors for extract the most useful information.

*Keywords*: Sensor Limitations, Remote Sensing, Spatial, Spectral, Resolution, Image Fusion.

## 1. INTRODUCTION

Remote sensing on board satellites techniques have proven to be powerful tools for the monitoring of the Earth's surface and atmosphere on a global, regional, and even local scale, by providing important coverage, mapping and classification of land cover features such as vegetation, soil, water and forests [1]. However, sensor limitations are most often a serious drawback since no single sensor offers at same time the optimal spectral, spatial and temporal resolution.

Many survey papers have been published recently, providing overviews of the history, developments, and the current state of the art of remote sensing data processing in the image-based application fields [2-4], but the major limitations in remote sensing fields has not been discussed in detail as well as image fusion methods. Only few researchers introduced that problems or limitations of image fusion which we can see in other section. The objectives of this paper are to present an overview of the major limitations in remote sensor satellite image and cover the multi-sensor image fusion. The paper is organized into six sections. Section 2 describes the Background upon Remote Sensing; under this section there are some other things like; remote sensing images; remote sensing Resolution Consideration; such as Spatial Resolution, spectral Resolution, Radiometric Resolution, temporal Resolution; data volume; and Satellite data with the resolution dilemma. Section 3 describes multi-sensors Images; there are sub sections like; processing levels of image fusion; categorization of image fusion techniques with our attitude towards categorization; Section 4 describes the discussion on the problems of available techniques. And the conclusions are drawn in Section 5.

## 2. Background upon Remote Sensing

The term "remote sensing" is most commonly used in connection with electromagnetic techniques of information acquisition [5]. These techniques cover the whole electromagnetic spectrum from low-frequency radio waves through the microwave, sub-millimeter, far infrared, near infrared, visible, ultraviolet, x-ray, and gamma-ray regions of the spectrum. Although this definition may appear quite abstract, most people have practiced a form of remote sensing in their lives. Remote sensing on board satellites techniques , as a science , deals with the acquisition , processing , analysis , interpretation , and utilization of data obtained from aerial and space platforms (i.e. aircrafts and satellites ) [6] .

In spaceborne remote sensing, sensors are mounted on-board a spacecraft orbiting the earth. There are several remote sensing satellites often launched into special orbits, geostationary orbits or sun synchronous orbits. In geostationary, the satellite will appear stationary with respect to the earth surface [7]. These orbits enable a satellite to always view the same area on the earth such as meteorological satellites. The earth observation satellites usually follow the sun synchronous orbits. A Sun synchronous orbit is a near polar orbit whose altitude is the one that the satellite will always pass over a location at given latitude at the same local time [7], such that (IRS, Landsat, SPOT…etc.).

There are two basic types of remote sensing system according to the source of energy: passive and active systems. A passive system (e.g. Landsat TM, SPOT-3 HRV) uses the sun as the source of electromagnetic radiation. Radiation from the sun interacts with the surface (for example by reflection) and the detectors aboard the remote sensing platform measure the amount of energy that is reflected. An active remote sensing system (e.g. on ERS-2 and RADAR-SAT) carries onboard its own electromagnetic radiation source. This electromagnetic radiation is directed to the surface and the energy that is reflected back from the surface is recorded [6] .This energy is associated with a wide range of wavelengths, forming the electromagnetic spectrum. Wavelength is generally measured in micrometers ($1 \times 10^{-6}$ m, µm). Discrete sets of continuous wavelengths (called wavebands) have been given names such as the microwave band, the infrared band, and the visible band. An example is given in Fig.1, which shows only a part of the overall electromagnetic spectrum. It is apparent that the visible waveband (0.4 to 0.7 µm), which is sensed by human eyes, occupies only a very small portion of the electromagnetic spectrum. A specific remote sensing instrument is designed to operate in one or more wavebands, which are chosen with the characteristics of the intended target in mind [8]. Those electromagnetic radiations pass through composition of the atmosphere to reach the Earth's surface features. The atmospheric constituents cause wavelength dependent absorption and scattering of radiation. The wavelengths at which electromagnetic radiation is partially or wholly transmitted through the atmosphere are known as atmospheric windowing [6]. The sensors on remote sensing systems must be designed in such a way as to obtain their data within these well- defined atmospheric windows. A significant research base has established the value of Remote Sensing for characterizing atmospheric; surface conditions; processes and these instruments prove to be one of the most cost effective means of recording quantitative information about our earth. In recent decades, the advent of satellite-based sensors has extended our ability to record information remotely to the entire earth and beyond. The electromagnetic spectrum proves to be so valuable because different portions of the electromagnetic spectrum react consistently to surface or atmospheric phenomena in specific and predictable ways. A single surface material will exhibit a variable response across the electromagnetic spectrum that is unique and is typically referred to as a spectral curve. Fig.2 provides an example of a typical electromagnetic spectrum response to green vegetation. Vegetation has a high reflectance in the near infrared band, while reflectance is lower in the red band. Knowledge of surface material Reflectance characteristics provide us with a principle based on which suitable wavebands to scan the Earth surface. A greater number of bands mean that more portions of the spectrum are recorded and greater discrimination can be applied to determining what a particular surface material or object is. Sensors that collect up to 16 bands of data are typically referred to as multispectral sensors while those that collect a greater number (typically up to 256) are referred to as hyperspectral.

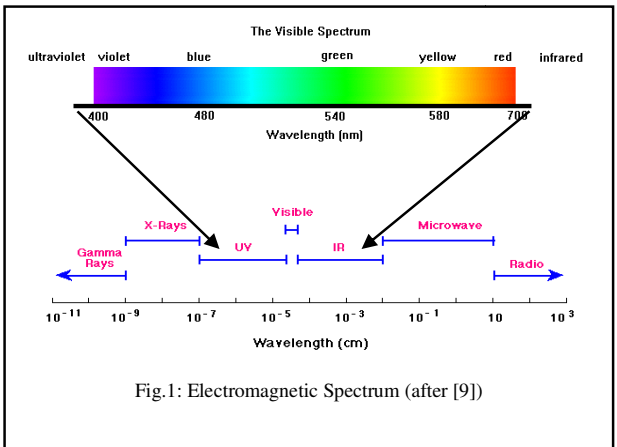

Fig.1: Electromagnetic Spectrum (after [9])

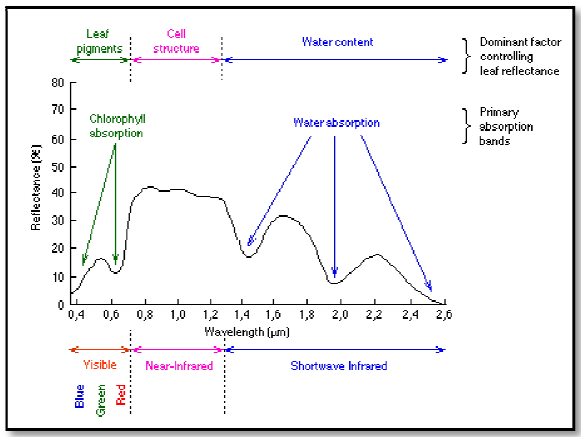

Fig.2: Spectral Curve for Green Vegetation (after [10])

Generally, remote sensing has become an important tool in many applications, which offers many advantages over other methods of data acquisition:

- Satellites give the spatial coverage of large areas and high spectral resolution.
- Satellites can view a given area repeatedly using the same imaging parameters.
- The digital data format of remote sensing allows direct digital processing of images and the integration with other data.

### 2.1 REMOTE SENSING IMAGES

Remote sensing images are available in two forms: photographic film form and digital form, which are related to a property of the object such as reflectance. When a collection of remotely sensed imagery and photographs considered, the general term "imagery" is often applied. An image is two types a monochrome image and a multicolour image. A monochrome image is a 2-dimensional light intensity function, $f(x, y)$ where x and y are spatial coordinates and the value of $f$ at $(x, y)$ is proportional to the brightness of the image at that point. If we have a multicolour image, $f$ is a vector, each component of which indicates the brightness of the image at point $(x, y)$ at the corresponding color band.

A digital image is an image $f(x, y)$ that has been discretized both in spatial co- ordinates and in brightness. It is represented by a 2-dimensional integer array, or a series of 2-dimensional arrays, one for each colour band [11]. The digitized brightness value is called the grey level value. Note that a digital image is composed of a finite number of elements, each of which has a particular location and value.

Each element is referred to as picture element, image element, pel, and pixel [12], even after defining it as a 'picture element'. Pixel can mean different things in different contexts and sometimes-conflicting contexts are present simultaneously. A pixel might be variously thought of [13]:

1. A single physical element of a sensor array. For example, the photosets on a semiconductor X-ray detector array or a digital camera sensor.
2. An element in an image matrix inside a computer. For $m \times n$ gray scale image there will be one $m \times n$ matrix. For $m \times n$ RGB color image there will be three $m \times n$ matrices, or one $m \times n \times 3$ matrix.
3. An element in the display on a monitor or data projector. As for the digital color sensor, each pixel of a color monitor display will comprise red, green and blue elements. There is rarely a one-to-one correspondence between the pixels in a digital image and the pixels in the monitor that displays the image.

The image data is rescaled by the computer's graphics card to display the image at a size and resolution that suits the viewer and the monitor hardware.

In remote sensing image, a Pixel is the term most widely used to denote the elements of a digital image. Each pixel represents an area on the Earth's surface. A pixel has an intensity value and a location address in the two dimensional image. The intensity value represents the measured physical quantity such as the solar radiance in a given wavelength band reflected from the ground, emitted infrared radiation or backscattered radar intensity. This value is normally the average value for the whole ground area covered by the pixel.

The intensity of a pixel digitized and recorded as a digital number. Due to the finite storage capacity, a digital number is stored with a finite number of bits (binary digits). Frequently the radiometric resolution expressed in terms of the number of binary digits, or bits, necessary to represent the range of available brightness values [18]. For example, an 8-bit digital number will range from 0 to 255 (i.e. $2^8$). The detected intensity value needs to scaled and quantized to fit within this range of value. In a radiometric calibrated image, the actual intensity value derived from the pixel digital number.

The field of digital image processing refers to processing digital images by means of a digital computer [14]. digital image processing has a broad spectrum of applications, such as remote sensing via satellites and other spacecrafts, image transmission and storage for business applications, medical processing, radar, sonar, and acoustic image processing, robotics, and automated inspection of industrial parts [15]. In order to extract useful information from the remote sensing images, Image Processing of remote sensing has been developed in response to three major problems concerned with pictures [11]:

- Picture digitization and coding to facilitate transmission, printing and storage of pictures.
- Picture enhancement and restoration in order, for example, to interpret more easily pictures of the surface of other planets taken by various probes.
- Picture segmentation and description as an early stage in Machine Vision.

There are different images for Interpretation corresponding to the images type such as; Multispectral and panchromatic (PAN) which consists of only one band and displayed as a gray scale image. A significant advantage of multi-spectral imagery is the ability to detect important differences between surface materials by combining spectral bands. Within a single band, different materials may appear virtually the same. By selecting particular band combination, various materials can be contrasted against their background by using colour. The colour composite images will display true colour or false colour composite images. The true colour of the resulting color composite image resembles closely to what the human eyes would observe. While the false colour occurs with composite the near or short infrared bands, the blue visible band is not used and the bands are shifted-visible green sensor band to the blue colour gun, visible red sensor band to the green colour gun and the NIR band to the red color gun.

## 2.2 REMOTE SENSING RESOLUTION CONSIDERATION

The Earth observation satellites offer a wide variety of image data with different characteristics in terms of spatial, spectral, radiometric, and temporal resolutions (see Fig.3). Resolution is defined as the ability of an entire remote-sensing system to render a sharply defined image. Resolution of a remote sensing is different types. Nature of each of these types of resolution must be understood in order to extract meaningful biophysical information from the remote sensed imagery [16].

### A. Spatial Resolution

The spatial resolution of an imaging system is not an easy concept to define. It can be measured in a number of different ways, depending on the user's purpose. The most commonly used measure, based on the geometric properties of the imaging system is the instantaneous field of view (IFOV) of sensor [17]. The IFOV is the ground area sensed by the sensor at a given instant in time. The spatial resolution is dependent on the IFOV. The dimension of the ground-projected is given by IFOV, which is dependent on the altitude and the viewing angle of sensor [6]. The finer the IFOV is, the higher the spatial resolution will be. However, this intrinsic resolution can often be degraded by other factors, which introduce blurring of the image, such as improper focusing, atmospheric scattering and target motion. Other methods of measuring the spatial resolving power of an imaging system based upon the ability of the system to distinguish between specified targets [17]. Then we can say that a spatial resolution is essentially a measure of the smallest features that can be observed on an image [6]. For instance, a spatial resolution of 79 meters is coarser than a spatial resolution of 10 meters. Generally, the better the spatial resolution is the greater the resolving power of the sensor system will be [6]. Other meaning of spatial resolution is the clarity of the high frequency detail information available in an image. Spatial resolution is usually expressed in meters in remote sensing and in document scanning or printing it is expressed as dots per inch (dpi).

### B. Spectral Resolution

Spectral resolution refers to the dimension and number of specific wavelength intervals in the electromagnetic spectrum to which a sensor is sensitive. For example, the SPOT panchromatic sensor is considered to have coarse spectral resolution because it records EMR between 0.51 and 0.73 µm. On the other hand, band 3 of the Landsat TM sensor has fine spectral resolution because it records EMR between 0.63 and 0.69 µm [16].

Generally, Spectral resolution describes the ability of a sensor to define fine wavelength intervals. The higher the spectral resolution is, the narrower the spectral bandwidth will be. If the platform has a few spectral bands, typically 4 to 7 bands, they are called multispectral, and if the number of spectral bands in hundreds, they are called hyperspectral data.

## C. Radiometric Resolution

The radiometric resolution of a remote sensing system is a measure of how many gray levels are measured between pure black and pure white [6]. The number of gray levels can be represented by a greyscale image is equal to $2^n$, where n is the number of bits in each pixel [20]. Frequently the radiometric resolution is expressed in terms of the number of binary digits, or bits necessary to represent the range of available brightness values [18, 20]. A larger dynamic range for a sensor results in more details being discernible in the image. The Landsat sensor records 8-bit images; thus, it can measure 256 unique gray values of the reflected energy while Ikonos-2 has an 11-bit radiometric resolution (2048 gray values). In other words, a higher radiometric resolution allows for simultaneous observation of high and low contrast objects in the scene [21].

## D. Temporal resolution

Temporal resolution refers to the length of time it takes for a satellite to complete one entire orbit cycle. It also refers to how often a sensor obtains imagery of a particular area. For example, the Landsat satellite can view the same area of the globe once every 16 days. SPOT, on the other hand, can revisit the same area every three days. Therefore, the absolute temporal resolution of a remote sensing system to image the exact same area at the same viewing angle a second time is equal to this period. Satellites not only offer the best chances of frequent data coverage but also of regular coverage.

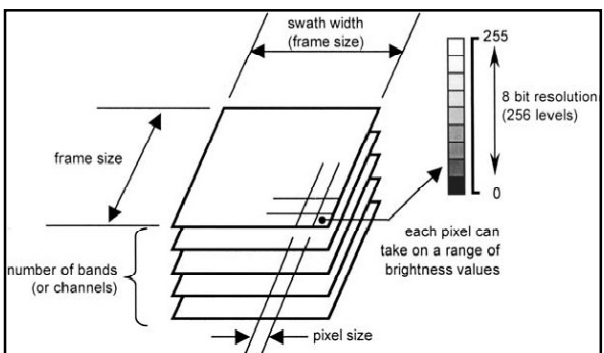

Fig.3: Technical characteristics of digital image data [18]

## E. Data Volume

The volume of the digital data can potentially be large for multi-spectral data, as a given area covered in many different wavelength bands. For example, a 3-band multi-spectral SPOT image covers an area of about $60 \times 60 \text{ km}^2$ on the ground with a pixel separation of 20m. So there are about $3000 \times 3000$ pixels per image, each pixel value in each band coded using an 8-bit (i.e. 1 byte) digital number, giving about 27 million bytes per image.

In comparison, the PAN data has only one band. Thus, PAN systems normally designed to give a higher spatial resolution than the multi-spectral system. For example, a SPOT PAN scene has the same coverage of about $60 \times 60 \text{ km}^2$ but the pixel size is 10 m, giving about $6000 \times 6000$ pixels and a total of about 36 million bytes per image. If a multi-spectral SPOT scene digitized also at 10 m pixel size, the data volume will be 108 million bytes.

## 2.3 SATELLITE DATA AND THE RESOLUTION DILEMMA

Current sensor technology allows the deployment of high resolution satellite sensors, but there are a major limitation of Satellite Data and the Resolution Dilemma as the fallowing:
- There is a tradeoff between spectral resolution and SNR.
- There is a tradeoff between radiometric resolution and SNR.
- There is a tradeoffs related to data volume and spatial resolution
- There is a tradeoff between the spatial and spectral resolutions.

For explain the above limitations as the following:

➢ The tradeoff between spectral resolution and SNR. Unfortunately, it is not possible to increase the spectral resolution of a sensor simply to suit the user's needs; there is a price to pay. Higher spectral resolution reduces the SNR of the sensor output. The signal is the information content of the data received at the sensor, while the noise is the unwanted variation that added to the signal. A compromise must be sought between the two in requirements of narrow band (high spectral resolution) and a low SNR [17].

➢ The tradeoff between radiometric resolution and SNR. There is no point in having a step size less than the noise level in the data. A low-quality instrument with a high noise level would necessary, therefore, have a lower radiometric resolution compared with a high-quality, high signal-to-noise-ratio instrument. Also higher radiometric resolution may conflict with data storage and transmission rates.

➢ In [22] described tradeoffs related to data volume and spatial resolution "the *increase in spatial resolution leads to an exponential increase in data quantity (which becomes particularly important when multispectral data should be collected). Since the amount of data collected by a sensor has to be balanced against the state capacity in transmission rates, archiving and processing capabilities. This leads to the dilemma of limited data volumes, an increase in spatial resolution must be compensated by a decrease in other data sensitive parameters, e.g. swath width, spectral and radiometric resolution, observation and data transmission duration*".

➢ Imaging sensors have a certain SNR based on their design. The energy reflected by the target must have a signal level large enough for the target to be detected by the sensor. The signal level of the reflected energy increases if the signal is collected over a larger IFOV or if it is collected over a broader spectral bandwidth. Collecting energy over a larger IFOV reduces the spatial resolution while collecting it over a larger bandwidth reduces its spectral resolution. Thus, there is a tradeoff between the spatial and spectral resolutions of the sensor [21].

Most, optical remote sensing satellites carry two types of sensors: the PAN and the MS sensors. The multispectral sensor records signals in narrow bands over a wide IFOV while the PAN sensor records signals over a narrower IFOV and over a broad range of the spectrum. Thus, the MS bands have a higher spectral resolution, but a lower spatial resolution compared to the associated PAN band, which has a higher spatial resolution and a lower spectral resolution [21]. Currently the spatial resolution of satellite images in optical remote sensing dramatically increased from tens of metres to metres and to < 1-metre (sees Table 1).

Table 1: Optical earth observation satellites, sensors, and their spatial and spectral resolutions [*23,24*].

| Optical satellite | Spatial resolution (m) (# of bands) | | | | Swath (km) | Year of launch |
|---|---|---|---|---|---|---|
| | Pan* | MS* | | | | |
| | | VNIR* | SWIR* | TIR* | | |
| Landsat 5 | | 30 (4) | 30 (2) | 120 (1) | 185 | 1984 |
| SPOT 2 | 10 | 20 (3) | | | 60 | 1990 |
| IRS-P2 | | 36.4 (4) | | | 74 | 1994 |
| IRS-1C | 5.8 | 23.5 (3) | 70.5 (1) | | 70, 142 | 1995 |
| IRS-1D | 5.8 | 23.5 (3) | 70.5 (1) | | 70, 142 | 1997 |
| SPOT 4 | 10 | 20 (3) | 20 (1) | | 60 | 1998 |
| Landsat 7 | 15 | 30 (4) | 30 (2) | 60 (1) | 185 | 1999 |
| CBERS 1 and 2 | 20 | 20 (4) | | | 113 | 1999, 2003 |
| Ikonos 2 | 1 | 4 (4) | | | 11 | 1999 |
| Terra/ASTER | | 15 (3) | 30 (6) | 90 (5) | 60 | 1999 |
| KOMPSAT-1 | 6.6 | | | | 17 | 1999 |
| EROS A1 | 1.9 | | | | 14 | 2000 |
| Quickbird 2 | 0.61 | 2.44 (4) | | | 16 | 2001 |
| SPOT 5 | 2.5-5 | 10 (3) | 20 (1) | | 60 | 2002 |
| IRS-P6 / ResourceSat-1 | 6 | 6(3), 23.5 (3) | | | 24, 70, 140 | 2003 |
| DMC-AlSat1 | | 32 (3) | | | 600 | 2002 |
| DMC-BILSAT-1 | 12 | 28 (4) | | | 25, 55 | 2003 |
| DMC-NigeriaSat 1 | | 32 (3) | | | 600 | 2003 |
| UK-DMC | | 32(3) | | | 600 | 2003 |
| OrbView-3 | 1 | 4 (4) | | | 8 | 2003 |
| DMC-Beijing-1 | 4 | 32 (3) | | | 24, 600 | 2005 |
| TopSat | 2.5 | 5 (3) | | | 25 | 2005 |
| KOMPSAT-2 | 1 | 4 (4) | | | 15 | 2006 |
| IRS-P5/CartoSat-1 | 2.5 | | | | 30 | 2006 |
| ALOS | 2.5 | 10(4) | | | 35, 70 | 2006 |
| Resurs DK-1 | 1 | 3 (3) | | | 28.3 | 2006 |
| WorldView-1 | 0.5 | | | | 17.5 | 2007 |
| RazakSat | 2.5 | 5 (4) | | | 20 | 2008 |
| RapidEye A–E | 6.5 | 6.5 (5) | | | 78 | 2008 |
| GeoEye-1 | 0.41 | 1.64 (4) | | | 15 | 2008 |
| EROS B – C | 0.7 | 2.8 | | | 16 | 2009 |
| WorldView-2 | 0.46 | 1.84 (8) | | | 16 | 2009 |
| Plèiades-1 and 2 | 0.7 | 2.8 (4) | | | 20 | 2010, 2011 |
| CBERS 3 and 4 | 5 | 20 (4), 40 | 40 (2) | 80 | 60, 120 | 2009, 2011 |

\* Pan: panchromatic; MS: multispectral; VNIR: visible and near infrared; SWIR: short wave infrared; TIR: thermal infrared.

Sensors all having a limited number of spectral bands (e.g. Ikonos and Quickbird) and there are only a few very high spectral resolution sensors with a low spatial resolution. But there is a trade-off in spectral and spatial resolution will remain.

3. **BACKGROUND FOR MULTISENSOR DATA FUSION**

Image fusion is a sub area of the more general topic of data fusion [25]. The concept of multi-sensor data fusion is hardly new while the concept of data fusion is not new [26]. The concept of data fusion goes back to the 1950's and 1960's, with the search for practical methods of merging images from various sensors to provide a composite image. This could be used to better identify natural and manmade objects [27]. Several words of fusion have appeared, such as merging, combination, synergy, integration. Different definitions can be found in literature on data fusion, each author interprets this term differently depending on his research interests. A general definition of data fusion is given by group set up of the European Association of Remote Sensing Laboratories (EARSeL) and the French Society for Electricity and Electronics (SEE, French affiliate of the IEEE), established a lexicon of terms of reference. Based upon the works of this group, the following definition is adopted and will be used in this study: "Data fusion is a formal framework which expresses means and tools for the alliance of data originating from different sources. It aims at obtaining information of greater quality; and the exact definition of 'greater quality' will depend upon the application" [28]. Image fusion forms a subgroup within this definition and aims at the generation of a single image from multiple image data for the extraction of information of higher quality. Having that in mind, the achievement of high spatial resolution, while maintaining the provided spectral resolution, falls exactly into this framework [29].

### 3.1 Processing Levels of Image Fusion

Multi-sensor data fusion can be performed at three different processing levels according to the stage at which fusion takes place i.e. pixel, feature and decision level of representation [29]. The following description and illustrations of fusion levels (see Fig.4) are given in more detail.

#### 3.1.1 Pixel Level Fusion

In pixel-level fusion, this is the lowest level of processing a new image formed through the combination of multiple images to increase the information content associated with each pixel. It uses the DN or radiance values of each pixel from different images in order to derive the useful information through some algorithms. An illustration is provided in Fig.4.a. The pixel based fusion of PAN and MS is also a pixel level fusion where new values are created or modelled from the DN values of PAN and MS images.

#### 3.1.2 Features of Level Fusions

This is an intermediate level image fusion. This level can be used as a means of creating additional composite features. Features can be pixel intensities or edge and texture features [30]. The Various kinds of features are considered depending on the nature of images and the application of the fused image. The features involve the extraction of feature primitives like edges, regions, shape, size, length or image segments, and features with similar intensity in the images to be fused from different types of images of the same geographic area. These extracted features are then combined using statistical approaches or other types of classifiers (see Fig.4.b).

It must be noted here that feature level fusion can involve fusing the feature sets of the same raw data or the feature sets of different sources of data that represent the same imaged scene. Also, if the feature sets originated from the same feature extraction or selection algorithm applied to the same data, the feature level fusion should be easy. However, feature level fusion is difficult to achieve when the feature sets are derived from different algorithms and data sources [31].

#### 3.1.3 Decision/Object Level Fusion

Decision-level fusion consists of merging information at a higher level of abstraction, combines the results from multiple algorithms to yield a final fused decision (see Fig.4.c). Input images are processed individually for information extraction. The obtained information is then combined applying decision rules to reinforce common interpretation [32].

### 3.2. Categorization of Image Fusion Techniques

There are many PAN sharpening techniques or Pixel-Based image fusion procedure techniques. There are two types of image fusion procedure available in literature:
1. Image Fusion Procedure Techniques Based on using the PAN Image.
2. Image Fusion Procedure Techniques Based on the Tools.

- In [22] Proposed the first type of categorization of image fusion techniques, depending on how the PAN information is used during the fusion procedure techniques, can be grouped into three classes: Fusion Procedures Using All Panchromatic Band Frequencies, Fusion Procedures Using Selected Panchromatic Band Frequencies and Fusion Procedures Using the Panchromatic Band Indirectly .

The second type of Image Fusion Procedure Techniques Based on the Tools found in many literatures different categorizations such as:
- In [33] classify PAN sharpening techniques into three

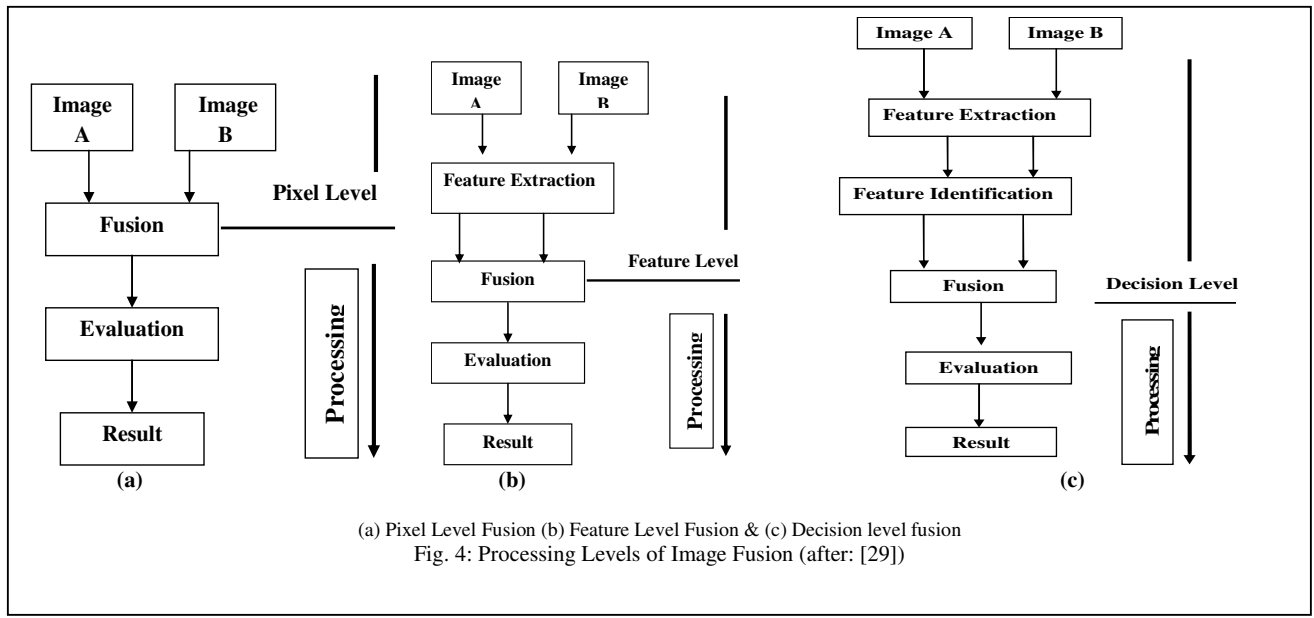

(a) Pixel Level Fusion (b) Feature Level Fusion & (c) Decision level fusion
Fig. 4: Processing Levels of Image Fusion (after: [29])

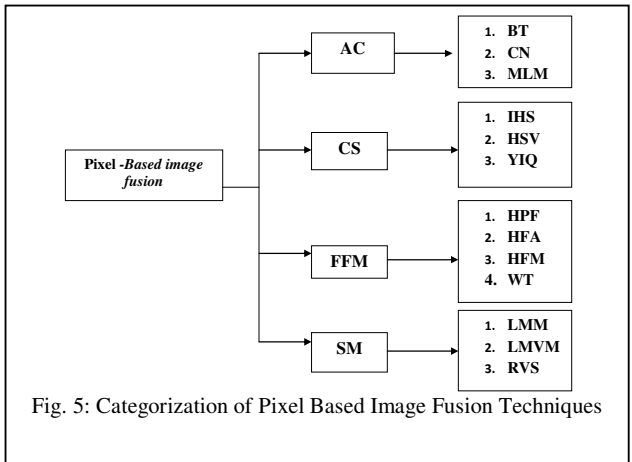

Fig. 5: Categorization of Pixel Based Image Fusion Techniques

classes: colour-related techniques, statistical methods and numerical methods. The first class includes colour compositions of three image bands in the RGB colour space as well as the more sophisticated colour transformations. The second class includes band statistics, such as the principal component (PC) transform. The third class includes arithmetic operations such as image multiplication, summation and image rationing as well as sophisticated numerical approaches such as wavelets.

- In [34] introduced another categorization of image fusion techniques: projection and substitution methods, relative spectral contribution and the 'spatial improvement by injection of structures' (ame´loration de la resolution spatial par injection de structures ARSIS) concept. In the first class are those methods, which project the image into another coordinate system and substitute one component. The second class is comparable with the second class of [33], with the exception that this category is restricted to band rationing and arithmetic combinations. The basis of the ARSIS concept is a multi-scale technique to inject the high spatial information into the multispectral images. Although this classification scheme bears some merits.

- In [35] classified the algorithms for pixel-level fusion of remote sensing images into three categories: the component substitution (CS) fusion technique, modulation-based fusion techniques and multi-resolution analysis (MRA)-based fusion techniques.

This work proposed another categorization scheme of image fusion techniques Pixel based image fusion methods because of its mathematical precision. It can be grouped into four categories based Fusion Techniques (Fig.5 shows the proposed categorization of pixel based image fusion Techniques):
1. Arithmetic Combination (AC)
2. Component Substitution (CS)
3. Frequency Filtering Methods (FFM)
4. Statistical Methods (SM)

### 3.2.1 The AC Methods

This category includes simple arithmetic techniques. Different arithmetic combinations have been employed for fusing MS and PAN images. They, directly, perform some type of arithmetic operation on the MS and PAN bands such as addition, multiplication, normalized division, ratios and subtraction which have been combined in different ways to achieve a better fusion effect. These models assume that there is high correlation between the PAN and each of the MS bands [32]. Some of the popular AC methods for pan sharpening are the Bovey Transform (BT); Colour Normalized Transformation (CN); Multiplicative Method (MLT) [36].

### 3.2.2 The CS Techniques

The methods under this category involve the transformation of the input MS images into new components. The transformation techniques in this class are based on the change of the actual colour space into another space and replacement of one of the new gained components by a more highly resolved image. The CS fusion techniques consist of three steps. First, forward transformation is applied to the MS bands after they have been registered to the PAN band. Second, one component of the new data space similar to the PAN band is replaced with the higher resolution band. Third, the fused results are constructed by means of inverse transformation to the original space [35].
Some of the popular CS methods for pan sharpening are the Intensity Hue Saturation IHS; Intensity Hue Value HSV; Hue Luminance Saturation HLS and Luminance I component (in-phase, an orange - cyan axis) Q component (Quadrature, a magenta - green axis) YIQ [37].

### 3.2.3 The FFM

Many authors have found fusion methods in the spatial domain (high frequency inserting procedures) superior over the other approaches, which are known to deliver fusion results that are spectrally distorted to some degree [38].

Fusion techniques in this group use high pass filters, Fourier transform or wavelet transform, to model the frequency components between the PAN and MS images by injecting spatial details in the PAN and introducing them into the MS image. Therefore, the original spectral information of the MS channels is not or only minimally affected [22]. Such algorithms make use of classical filter techniques in the spatial domain. Some of the popular FFM for pan sharpening are the High-Pass Filter Additive Method (HPFA) [39-40], High –Frequency- Addition Method (HFA)[36] , High Frequency Modulation Method (HFM) [36] and The Wavelet transform-based fusion method (WT) [41-42].

### 3.2.4 Statistical Methods (SM) based image Fusion

Different SM have been employed for fusing MS and PAN images. They perform some type of statistical variable on the MS and PAN bands. The SM used to solve the two major problems in image fusion colour distortion and operator (or dataset) dependency. It is different from pervious image fusion techniques in two principle ways: It utilizes the statistical variable such as the least squares;

average of the local correlation or the variance with the average of the local correlation techniques to find the best fit between the grey values of the image bands being fused and to adjust the contribution of individual bands to the fusion result to reduce the colour distortion. Some of the popular SM methods for pan sharpening are Local Mean Matching (LMM), Local Mean and Variance Matching (LMVM), Regression variable substitution (RVS), and Local Correlation Modelling (LCM) [43-44].

## 4. DISCUSSION ON PROBLEMS OF AVAILABLE TECHNIQUES

The available fusion techniques have many limitations and problems. The Problems and limitations associated with these fusion techniques which reported by many studies [45-49] as the following:

- The most significant problem is the colour distortion of fused images. A major reason for the insufficiency of available techniques fusion is the change of the PAN spectral range. The wavelength of the PAN image is much broader than multispectral bands. This discrepancy between the wavelengths causes considerable colour distortion to occur when fusing high resolution PAN and MS images.
- Most of the existing methods were developed for the fusion of "low" spatial resolution images such as SPOT and Land-sat TM they may or may not be suitable for the fusion of VHR image for specific tasks. In addition, operator dependency was also a main problem of existing fusion techniques, i.e. different operators with different knowledge and experience usually produced different fusion results for same method.

- The jury is still out on the benefits of a fused image compared to its original images. There is also a lack of measures for assessing the objective quality of the spatial and spectral resolution for the fusion methods.

## 5. CONCLUSION

Briefly, one can conclude that improving a satellite sensor's resolution may only be achieved at the cost of losing some original advantages of satellite remote sensing. Due to the underlying physics principles, therefore, it is usually not possible to have both very high spectral and spatial resolution simultaneously in the same remotely sensed data especially from orbital sensors, with the fast development of modern sensor technologies however, technologies for effective use of the useful information from the data are still very limited. These limitations have significantly limited the effectiveness of many applications of satellite images required both spectral and spatial resolution to be high.

Therefore, multiple sensor data fusion introduced to solve these problems. However, Problems and limitations associated with them which explained in above section. Therefore, the research on technology development is necessary new advanced data fusion approaches are needed to improve the data of remote sensing. Also, a new performance assessment criteria and automatic quality assessment methods to evaluate the possible benefits of fusion and make final conclusions can be drawn on the most suitable method of fusion to make effectively use of these sensors.

**REFERENCES**

[1] Simone, G.; Farina, A.; Morabito, F.C.; Serpico, S.B;Bruzzone, L. .,2002. "Image fusion techniques for remote sensing applications". Inf. Fusion 2002, 3, 3–15.

[2] Myint, S.W., Yuan, M., Cerveny, R.S., Giri, C.P., 2008. Comparison of remote sensing image processing techniques to identify tornado damage areas from landsat TM data. Sensors 8 (2), pp.1128-1156.

[3] T. Blaschke, 2010. "Object based image analysis for remote sensing". Review Springer, ISPRS Journal of Photogrammetry and Remote Sensing 65 (2010) ,PP. 2-16.

[4] Kai Wang, Steven E. Franklin , Xulin Guo, Marc Cattet ,2010. "Remote Sensing of Ecology, Biodiversity and Conservation: A Review from the Perspective of Remote Sensing Specialists". Review article, Sensors 2010, 10, 9647-9667; doi:10.3390/s101109647

[5] Elachi C. and van Zyl J., 2006. "Introduction to the Physics and Techniques of Remote Sensing". John Wiley & Sons, Inc.

[6] Gibson P. J., 2000.Introductory Remote Sensing: Principles and Concepts. Routledge -Taylar & Francis Group.

[7] Roddy D., 2001."Satellite Communications".3rd Edition, McGraw-Hill Companies, Inc.

[8] Tso B. and Mather P. M., 2009. "Classification Methods For Remotely Sensed Data". Second Edition, Taylor & Francis Group, LLC.

[9] Saxby, G., 2002. "The Science of Imaging". Institute of Physics Publishing Inc., London.

[10] Hoffer, A.M., 1978. Biological and physical considerations in applying computeraided analysis techniques to remote sensor data, in Remote Sensing: The Quantitative Approach, P.H. Swain and S.M. Davis (Eds), McGraw-Hill Book Company, pp.227-289.

[11] Petrou M., 1999. "Image Processing The Fundamentals". Wiley & Sons,Ltd.

[12] Gonzalez R. C. and Woods R. E., 2002. "Digital Image Processing". Second Edition.Prentice-Hall, Inc.

[13] Bourne R., 2010. "Fundamentals of Digital Imaging in Medicine". Springer-Verlag London Ltd.

[14] Gonzalez R. C., Woods R. E. and Eddins S. L., 2004. "Digital Image Processing Using MATLAB®". Pearson Prentice-Hall.

[15] Jain A. K., 1989."Fundamentals of Digital Image Processing".Prentice-Hall,Inc.

[16] Jensen J.R., 1986. "Introductory Digital Image Processing A Remote Sensing Perspective". Englewood Cliffs, New Jersey: Prentice-Hall.

[17] Mather P. M., 1987. "Computer processing of Remotely Sensed Images". John Wiley & Sons.

[18] Richards J. A., and Jia X., 1999. "Remote Sensing Digital Image Analysis". 3rd Edition. Springer - verlag Berlin Heidelberg New York.

[19] Logan S., 1998. "Devloping Imaging Applications with XIELIB". Prentic Hall.

[20] Umbaugh S. E., 1998. "Computer Vision and Image Processing: Apractical Approach Using CVIP tools". Prentic Hall.

[21] Pradham P., Younan N. H. and King R. L., 2008. Concepts of image fusion in remote sensing applications". In Tania Stathaki "Image Fusion: Algorithms and Applications". Elsevier Ltd.pp.393-482.

[22] Hill J., Diemer C., Stöver O., Udelhoven Th.,1999. "A Local Correlation Approach For The Fusion Of Remote Sensing Data With Different Spatial Resolutions In Forestry Applications". International Archives

of Photogrammetry and Remote Sensing, Vol. 32, Part 7-4-3 W6, Valladolid, Spain, 3-4 June, 1999.

[23] Zhang Y.,2010. "Ten Years Of Technology Advancement In Remote Sensing And The Research In The CRC-AGIP Lab In GGE". GEOMATICA Vol. 64, No. 2, 2010 pp. 173 to 189.

[24] Stoney, W.E. 2008. ASPRS guide to land imaging satellites. http://www.asprs.org/news/satellites/ASPRS_DATA-BASE _021208.pdf [Last accessed Jan 15, 2012].

[25] Hsu S. H., Gau P. W., I-Lin Wu I., and Jeng J. H., 2009,"Region-Based Image Fusion with Artificial Neural Network". World Academy of Science, Engineering and Technology, 53, pp 156 -159.

[26] Llinas J.and Hall D. L., 1998, "An introduction to multi-sensor data fusion". IEEE, VI, N° 1, pp. 537-540.

[27] Wang Z., Djemel Ziou, Costas Armenakis, Deren Li, and Qingquan Li,2005..A Comparative Analysis of Image Fusion Methods. IEEE Transactions On Geoscience And Remote Sensing, Vol. 43, No. 6, JUNE 2005,pp. 1391-1402.

[28] Ranchin T. and Wald L., 2000. "Fusion of high spatial and spectral resolution images: the ARSIS concept and its implementation". Photogrammetric Engineering and Remote Sensing, Vol.66, No.1, pp.49-61.

[29] Pohl C., 1999." Tools And Methods For Fusion Of Images Of Different Spatial Resolution". International Archives of Photogrammetry and Remote Sensing, Vol. 32, Part 7-4-3 W6, Valladolid, Spain, 3-4 June,

[30] Kor S. and Tiwary U.,2004.'' Feature Level Fusion Of Multimodal Medical Images In Lifting Wavelet Transform Domain".Proceedings of the 26th Annual International Conference of the IEEE EMBS San Francisco, CA, USA, pp. 1479-1482

[31] Chitroub S., 2010. "Classifier combination and score level fusion: concepts and practical aspects". International Journal of Image and Data Fusion, Vol. 1, No. 2, June 2010, pp. 113–135.

[32] Dong J.,Zhuang D., Huang Y.,Jingying Fu,2009. "Advances In Multi-Sensor Data Fusion: Algorithms And Applications ". *Review* ,ISSN 1424-8220 Sensors 2009, 9, pp.7771-7784.

[33] Pohl C., Van Genderen J. L., 1998, "Multisensor image fusion in remote sensing: concepts, methods and applications", ".(Review Article), International Journal of Remote Sensing, Vol. 19, No. 5, pp. 823-854.

[34] Wald L., 1999, "Definitions And Terms Of Reference In Data Fusion". International Archives of Photogrammetry and Remote Sensing, Vol. 32, Part 7-4-3 W6, Valladolid, Spain, 3-4 June,

[35] Zhang J., 2010. "Multi-source remote sensing data fusion: status and trends", International Journal of Image and Data Fusion, Vol. 1, No. 1, pp. 5–24.

[36] Firouz A. Al-Wassai, N.V. Kalyankar , A. A. Al-Zuky, 2011. "Arithmetic and Frequency Filtering Methods of Pixel-Based Image Fusion Techniques ".IJCSI International Journal of Computer Science Issues, Vol. 8, Issue 3, No. 1, May 2011, pp. 113- 122.

[37] Firouz A. Al-Wassai, N.V. Kalyankar, A. A. Al-zuky ,2011. "The IHS Transformations Based Image Fusion". Journal of Global Research in Computer Science, Volume 2, No. 5, May 2011, pp. 70 – 77.

[38] Gangkofner U. G., P. S. Pradhan, and D. W. Holcomb, 2008. "Optimizing the High-Pass Filter Addition Technique for Image Fusion". Photogrammetric Engineering & Remote Sensing, Vol. 74, No. 9, pp. 1107–1118.

[39] Lillesand T., and Kiefer R.1994. "Remote Sensing And Image Interpretation". 3rd Edition, John Wiley And Sons Inc.

[40] Aiazzi B., S. Baronti , M. Selva,2008. "Image fusion through multiresolution oversampled decompositions". in Image Fusion: Algorithms and Applications ".Edited by: Stathaki T. "Image Fusion: Algorithms and Applications". 2008 Elsevier Ltd.

[41] Aiazzi, B., Baronti, S., and Selva, M., 2007. "Improving component substitution pan-sharpening through multivariate regression of MS+Pan data". IEEE Transactions on Geoscience and Remote Sensing, Vol.45, No.10, pp. 3230–3239.

[42] Malik N. H., S. Asif M. Gilani, Anwaar-ul-Haq, 2008. "Wavelet Based Exposure Fusion". Proceedings of the World Congress on Engineering 2008 Vol I WCE 2008, July 2 - 4, 2008, London, U.K.

[43] Firouz A. Al-Wassai, N.V. Kalyankar , A.A. Al-Zuky, 2011c." The Statistical methods of Pixel-Based Image Fusion Techniques". International Journal of Artificial Intelligence and Knowledge Discovery Vol.1, Issue 3, July, 2011 5, pp. 5- 14.

[44] Firouz A. Al-Wassai, N.V. Kalyankar, A. A. Al-zuky ,2011. " Multisensor Images Fusion Based on Feature-Level". International Journal of Advanced Research in Computer Science, Volume 2, No. 4, July-August 2011, pp. 354 – 362.

[45] Zhang J., 2010. "Multi-source remote sensing data fusion: status and trends", International Journal of Image and Data Fusion, Vol. 1, No. 1, pp. 5–24.

[46] Firouz A. Al-Wassai, N.V. Kalyankar, A. A. Al-zuky ,2011. " Multisensor Images Fusion Based on Feature-Level". International Journal of Advanced Research in Computer Science, Volume 2, No. 4, July-August 2011, pp. 354 – 362.

[47] Firouz Abdullah Al-Wassai, N.V. Kalyankar, 1012. "A Novel Metric Approach Evaluation for the Spatial Enhancement of Pan-Sharpened Images". Computer Science & Information Technology (CS & IT), 2(3), 479 – 493.

[48] Firouz Abdullah Al-Wassai, N.V. Kalyankar, Ali A. Al-Zaky, "Spatial and Spectral Quality Evaluation Based on Edges Regions of Satellite: Image Fusion", IEEE Computer Society, 2012 Second International Conference on Advanced Computing & Communication Technologies, ACCT 2012, pp.265-275.

[49] Firouz Abdullah Al-Wassai, N.V. Kalyankar, Ali A. Al-Zaky, "Spatial and Spectral Quality Evaluation Based on Edges Regions of Satellite: Image Fusion," ACCT, 2nd International Conference on Advanced Computing & Communication Technologies, 2012, pp.265-275.

**AUTHORS**

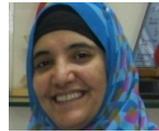

*Firouz Abdullah Al-Wassai*. Received the B.Sc. degree in, Physics from University of Sana'a, Yemen, Sana'a, in 1993. The M.Sc.degree in, Physics from Bagdad University , Iraqe, in 2003, Research student.Ph.D in the department of computer science (S.R.T.M.U), India, Nanded.

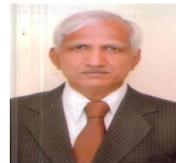

*Dr. N.V. Kalyankar*, Principal,Yeshwant Mahvidyalaya, Nanded(India) completed M.Sc.(Physics) from Dr. B.A.M.U, Aurangabad. In 1980 he joined as a leturer in department of physics at Yeshwant Mahavidyalaya, Nanded. In 1984 he completed his DHE. He completed his Ph.D. from Dr.B.A.M.U. Aurangabad in 1995. From 2003 he is working as a Principal to till date in Yeshwant Mahavidyalaya, Nanded. He is also research guide for Physics and Computer Science in S.R.T.M.U, Nanded. 03 research students are successfully awarded Ph.D in Computer Science under his guidance. 12 research students are successfully awarded M.Phil in Computer Science under his guidance He is also worked on various boides in S.R.T.M.U, Nanded. He is also worked on various bodies is S.R.T.M.U, Nanded. He also published 30 research papers in various international/national journals. He is peer team member of NAAC (National Assessment and Accreditation Council, India ). He published a book entilteld "DBMS concepts and programming in Foxpro". He also get various educational wards in which "Best Principal" award from S.R.T.M.U, Nanded in 2009 and "Best Teacher" award from Govt. of Maharashtra, India in 2010. He is life member of Indian "Fellowship of Linnean Society of London(F.L.S.)" on 11 National Congress, Kolkata (India). He is also honored with November 2009.